\documentclass[journal]{IEEEtran}

\usepackage{graphicx}
\usepackage{amsmath}
\allowdisplaybreaks[4]
\usepackage{array}
\usepackage{epsfig}
\usepackage{booktabs}
\usepackage{amssymb}
\usepackage{color}
\usepackage{array}
\usepackage{bm}
\usepackage{url}
\usepackage{multirow}
\usepackage{adjustbox}
\usepackage{graphicx}
\UseRawInputEncoding
\usepackage{booktabs}
\usepackage{siunitx}
\usepackage{multicol}
\usepackage{xcolor}
\definecolor{lightblue}{RGB}{180,199,231}
\definecolor{lightorange}{RGB}{244,177,131}
\definecolor{green}{RGB}{71,161,39}
\definecolor{gold}{RGB}{255,192,0}
\usepackage{pgfplots}
\usepackage{pgf-pie}
\usepackage[
CJKbookmarks=true, bookmarksnumbered=true, bookmarksopen=true,
colorlinks=true, 
pdfborder=001, 
citecolor=blue, linkcolor=blue, anchorcolor=red, urlcolor=black
]{hyperref}

\usepackage[tight,footnotesize]{subfigure}

\begin{document}\sloppy

\title{Masked Attribute Description Embedding for Cloth-Changing Person Re-identification}

\author{Chunlei Peng,~\IEEEmembership{Member,~IEEE,}
        Boyu Wang,
        Decheng Liu,
        Nannan Wang,~\IEEEmembership{Senior Member,~IEEE,}
        Ruimin Hu,
        and Xinbo~Gao,~\IEEEmembership{Fellow,~IEEE,}

\thanks{
C. Peng, B. Wang, and D. Liu are with the State Key Laboratory of Integrated Services Networks, School of Cyber Engineering, Xidian University, Xi'an 710071, Shaanxi, P. R. China, and with the Key Laboratory of Artificial Intelligence, Ministry of Education, Shanghai, 200240, China. (e-mail: clpeng@xidian.edu.cn; byw.xidian@gmail.com; dchliu@xidian.edu.cn).

N. Wang is with the State Key Laboratory of Integrated Services Networks, School of Telecommunications Engineering, Xidian University, Xi'an 710071, Shaanxi, P. R. China (e-mail: nnwang@xidian.edu.cn).

R. Hu is with the School of Cyber Engineering, Xidian University, Xi'an 710071, Shaanxi, P. R. China (e-mail: rmhu@xidian.edu.cn). 

X. Gao is with the Chongqing Key Laboratory of Image Cognition, Chongqing University of Posts and Telecommunications, Chongqing 400065, P. R. China (e-mail: gaoxb@cqupt.edu.cn).

Corresponding author: Nannan Wang.
}}


\maketitle

\begin{abstract}
Cloth-changing person re-identification (CC-ReID) aims to match persons who change clothes over long periods. The key challenge in CC-ReID is to extract cloth-irrelated features, such as face, hairstyle, body shape, and gait. Current research mainly focuses on modeling body shape using multi-modal biological features (such as silhouettes and sketches). However, it does not fully leverage the personal description information hidden in the original RGB image. Considering that there are certain attribute descriptions that remain unchanged after the changing of cloth, we propose a Masked Attribute Description Embedding (MADE) method that unifies personal visual appearance and attribute description for CC-ReID. Specifically, handling variable \textcolor{black}{cloth-sensitive} information, such as color and type, is challenging for effective modeling. To address this, we mask the \textcolor{black}{clothes type and color information (upper body type, upper body color, lower body type, and lower body color)} in the personal attribute description extracted through an attribute detection model. The masked attribute description is then connected and embedded into Transformer blocks at various levels, fusing it with the low-level to high-level features of the image. This approach compels the model to discard cloth information. Experiments are conducted on several CC-ReID benchmarks, including PRCC, LTCC, Celeb-reID-light, and LaST. Results demonstrate that MADE effectively utilizes attribute description, enhancing cloth-changing person re-identification performance, and compares favorably with state-of-the-art methods. The code is available at \href{https://github.com/moon-wh/MADE}{https://github.com/moon-wh/MADE}.
\end{abstract}

\IEEEpeerreviewmaketitle
\section{Introduction}
\label{sec:intro}

{\textcolor{black}{Cloth-changing person re-identification (CC-ReID) aims to match persons who change clothes over long periods. Traditional person re-identification (Re-ID) operates} under the assumption that the person being tracked moves within a confined area and time and will not change their clothes. However, in practical scenarios, persons captured by surveillance cameras may traverse larger areas and be observed over extended periods, during which they might change their \textcolor{black}{clothes}. This deviation in appearance challenges the reliability of color-based information utilized by earlier Re-ID approaches for person re-identification. Therefore, in recent years, cloth-changing person re-identification has attracted more and more attention.

In cloth-changing person re-identification, the reliance on color information by traditional methods becomes unreliable, and addressing the modeling of clothes changes poses a significant challenge~\cite{Alpher01}. Therefore, the key to solving CC-ReID lies in identifying information about personal \textcolor{black}{features} that is insensitive to these \textcolor{black}{clothes} changes. To mitigate the interference caused by varying clothes and uncover invariant features of persons, some cloth-changing person re-identification methods focus on the multi-modal biometric features of persons. PRCC~\cite{Alpher01} and FSAM~\cite{FSAM} study silhouette information of persons. 3DSL~\cite{3DSL} aims to learn the 3D shape features of persons. However, contours and 3D features eliminate all color information of images. MBUNet~\cite{MBUNet} contains a branch for extracting posture features. GI-ReID~\cite{GI-ReID} and ViT-VIBE~\cite{ViT-VIBE} utilize gait features. However, extracting pose features from a single image of a person is still a challenging task. SpTSkM~\cite{SpTSkM} uses skeleton normalization to assist person recognition. RF-ReID~\cite{RF-ReID} infers personal skeleton features from radio frequency signals. However, skeleton features are challenging to extract and utilize directly. \textcolor{black}{CAL~\cite{CAL} and the method proposed by Chan \emph{et al.}~\cite{ACMMM23} do not utilize multi-modal biometric features; instead, they use GAN~\cite{GAN} networks to extract cloth-irrelated features from pedestrian images. However, GAN networks have unstable training, long training times, and difficulty tuning hyperparameters. These methods typically require additional complex models to extract biometric features such as contours, 3D shapes, postures, and skeletons, which demand substantial computing resources for training and extraction and also require complex fusion of the extracted biometric features with image features. Methods that do not use multi-modal features often face issues with unstable training.}

\begin{figure*}[t]
	\centering
	\includegraphics[width=0.8\linewidth]{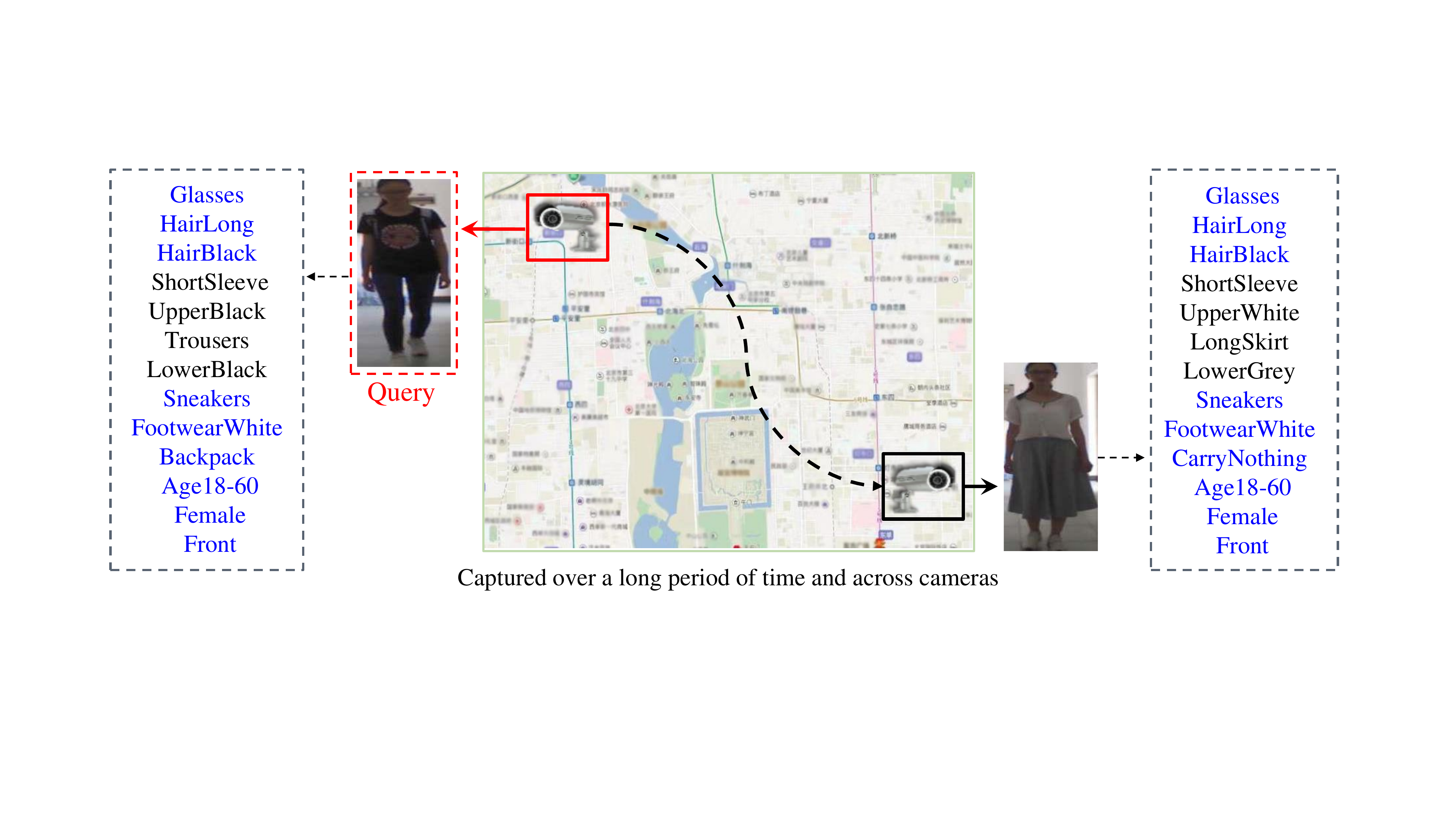}
	\caption{\color{black} An illustration of cloth-changing person re-identification over a long period of time and across cameras. The attributes of the person image is shown in the figure. Attributes related to clothes are marked in black, while attributes irrelevant to clothes are marked in blue. In the cloth-changing person re-identification scenario, many attributes unrelated to clothes remain consistent, such as hair, glasses, shoes, age, and gender, which could be useful for re-identification.}
	\label{fig:att}
\end{figure*}

In the context of cloth-changing person re-identification, even over extended time intervals, persons tend to alter only their clothes choices, while other attributes such as gender, age, and hair color remain consistent, as depicted in Fig. \ref{fig:att}. This \textcolor {black}{cloth-irrelevant}  attributes are helpful to person re-identification, while cloth-related attributes can be easily eliminated. This paper introduces the Masked Attribute Description Embedding (MADE) method to effectively mine cloth-irrelevant information from original RGB images of persons. Specifically, we adopt the attribute detection model \textcolor{black}{SOLIDER~\cite{SOLIDER}} to extract pedestrian attributes. \textcolor{black}{SOLIDER is a self-supervised learning framework used to learn universal human representations from a large number of unannotated human images. It demonstrates excellent performance on pedestrian attribute recognition tasks.} Then, by masking the cloth-related pedestrian attributes to obtain masked attribute descriptions (the definition of cloth-related attributes is in section \ref{3.2}), and the cloth-sensitive features are eliminated by shielding the cloth information in the RGB image and retaining other cloth-insensitive color features. Due to the editable nature of text descriptions, clothes color can be quickly and efficiently eliminated. MADE then connects and embeds the masked attributes description data encoded by Linear Projection into TrV blocks~\cite{EVA02} at different levels to fuse the image features. By mapping different feature spaces to a shared latent space, masked description can be fused with image features without the need for additional text encoders, forcing the model to discard cloth-sensitive information.

We summarise the contributions of this work as follows.
\begin{enumerate}
		\item We propose a Masked Attribute Description Embedding (MADE) re-identification method for cloth-changing person re-identification, which unifies the person's variable color visual appearance and editable attribute description in CC-ReID.
	
	\item \textcolor{black}{We introduce multi-modal attribute description information in CC-ReID, which is easier to extract and edit than skeletons or contours. By masking clothes and cloth-color items in these descriptions and embedding them into image features, the model is compelled to discard cloth-sensitive features.}
	
	\item \textcolor{black}{For the first time, we employ a simple, efficient method to integrate image features with attribute \textcolor{black}{descriptions} in CC-ReID. This approach maps descriptions and image features to a shared latent space, effectively allowing the model to capture their associations without additional text encoders.}
	
	\item Our extensive experiments on four public benchmark datasets, PRCC, LTCC,  Celeb-ReID-light, and LaST, show that MEDA consistently outperforms existing state-of-the-art methods by a large margin.
\end{enumerate}

In the following, we will discuss related work in section \ref{sec:Relat}. We will present the details of our proposed method in section \ref{sec:Method}. The experimental results are provided in section \ref{sec:exp}. Section \ref{sec:con} concludes this paper with our future research directions

\section{Related Work}
\label{sec:Relat}

\subsection{Multi-Modal Features based Cloth-Changing Person Re-Identification}

The core of solving cloth-changing person re-identification is to extract the cloth-irrelevant features in person images. To this end, some research focuses on multi-modal features that are less variable than clothes, such as silhouette, 3D shape, skeleton, walking posture, etc. 

FSAM~\cite{FSAM} uses a parsing network to train and obtain contour images, enabling coarse-to-fine mask learning.~\cite{MAC-DIM} proposes a multi-scale appearance and contour depth Infomax (MAC-DIM) to maximize the mutual information between appearance and contour shape features. Mu \emph{et al.}~\cite{Muetal} utilize human parsing models to segment the semantic parts of the human body to obtain the binary body shape masks. These methods discard all color information in the original RGB image in the contour processing module. However, some color information is helpful for person re-identification.

SpTSkM~\cite{SpTSkM} explores \textcolor{black}{personal} motion pattern information from 3D skeletons normalized by ST-GAN~\cite{ST-GAN} to assist person re-identification. 3DSL~\cite{3DSL} distinguishes different identities by learning 3D shape features and 3D reconstruction subnetworks. These methods obtain 3D shapes through cumbersome 3D parsing and processing networks, which increases the complexity of model training.

CESD~\cite{CESD} uses a pose detector to detect personal body joint points and uses shape embedding to separate clothes and distinguish shape information through joint point features. The pose feature branch of MBUNet ~\cite{MBUNet} applies the direction adaptive graph convolution layer to obtain the relevant information between different keypoints in heatmaps. ViT-VIBE~\cite{ViT-VIBE} uses ViT~\cite{ViT} to combine appearance and gait features learned through VIBE~\cite{VIBE}. However, these methods do not fully exploit and utilize the cloth-irrelevant features in the original image. 

In this paper, we propose MADE to unify \textcolor{black}{personal} appearance and language description. Multi-modal attribute description information is introduced in CC-ReID, which is more obvious and accessible to extract and edit than biological features such as skeleton or silhouette in original person images. Information that helps identify persons can be retained to the greatest extent while accurately removing interference from cloth information.


\subsection{Text-to-Image Person Re-Identification} 
Text-to-image person re-identification aims to search for pedestrian images of an interested identity via textual descriptions. The main challenge in this field is how to efficiently fuse image and text features into a joint embedding space. Early research work~\cite{Imge-text1, CUHK-PEDES} adopted VGG~\cite{VGG} and LSTM~\cite{LSTM} to learn the representation of visual-text modalities. CFine~\cite{CFine} proposes a CLIP-driven fine-grained information excavation framework to fully utilize the powerful knowledge of CLIP for text-image person re-identification. IRRA~\cite{IRRA} integrates visual cues with CLIP~\cite{Clip} encoded text tokens into a cross-modal multi-modal interaction encoder, enabling cross-modal interaction. These methods utilize sentence captions describing persons. However, when dealing with scenes involving changes in clothes, the model faces challenges in directly processing cloth-related fragments within the captions. Introducing an encoder to encode language text would increase the complexity of model training. To address this, we leverage itemized attributes. This approach enables the model to precisely handle cloth-related information within the description, avoiding the processing of the entire statement as a whole. Simultaneously, we embed the attribute vector directly into the Transformer block, eliminating the need for additional encoder encoding.

\subsection{Attribute-based Person Re-identification} 
It has been well exploited to perform person re-identification with attributes. APR~\cite{APR} introduces the Attribute Reweighting Module (ARM), which corrects predictions of attributes based on learned dependencies and correlations between attributes. AAB~\cite{AAB} utilizes fine-grained attribute attention modules to enhance the performance of the Re-ID task. MSPA~\cite{MSPA} uses ConvLSTM to memorize the relationship between \textcolor{black}{personal} attribute features. AMNet~\cite{AMNet} designs the Spatial Channel Attention Module (SCAM) to extract features from each attribute. Additionally, it utilizes the semantic reasoning and information propagation capabilities of graph convolutional networks to explore the relationship between attribute features and pedestrian features. UCAD~\cite{UCAD} proposes a clothes attribute decomposition network that can effectively attenuate the influence of clothes through loss function constraints. These methods utilize all pedestrian attributes to address person re-identification challenges but overlook the editability of the attributes description. We introduce the editability of description in CC-ReID, which can accurately remove cloth-sensitive attributes and help the model learn cloth-sensitive information.

\section{Method}
\label{sec:Method}

\subsection{Overview}

The key to achieving cloth-changing person re-identification lies in extracting cloth-insensitive features from the image. In CC-ReID, we introduce attribute description to mitigate the impact of clothes interference. Consequently, our model primarily focuses on modeling the relationship between and within the two types of images and descriptions. EVA-02~\cite{EVA02} is pre-trained to reconstruct powerful and robust language-aligned visual features through occlusion image modeling, resulting in transferable models. Based on this model, we proposed MADE framework to integrate personal masked attribute description data with image visual features, addressing the challenges of cloth-changing person re-identification. The framework of our approach is shown in Fig. \ref{fig:method}. Given an image sample $x_{i}$, $x_{i}$ extracts an editable attribute description through Description Extraction and Mask module. After the masked attribute description is converted into a binary vector, it is connected and embedded at different levels through Linear Projection to fuse with image features in TrV blocks~\cite{EVA02}. We introduce how to extract and mask attribute description in section \ref{3.2}. Then, section \ref{3.3} details how to add masked attribute description data to improve the performance of CC-ReID. Finally, we elaborate on the model's loss function and inference process.

\begin{figure*}[t]
	\centering
	\includegraphics[width=0.8\linewidth]{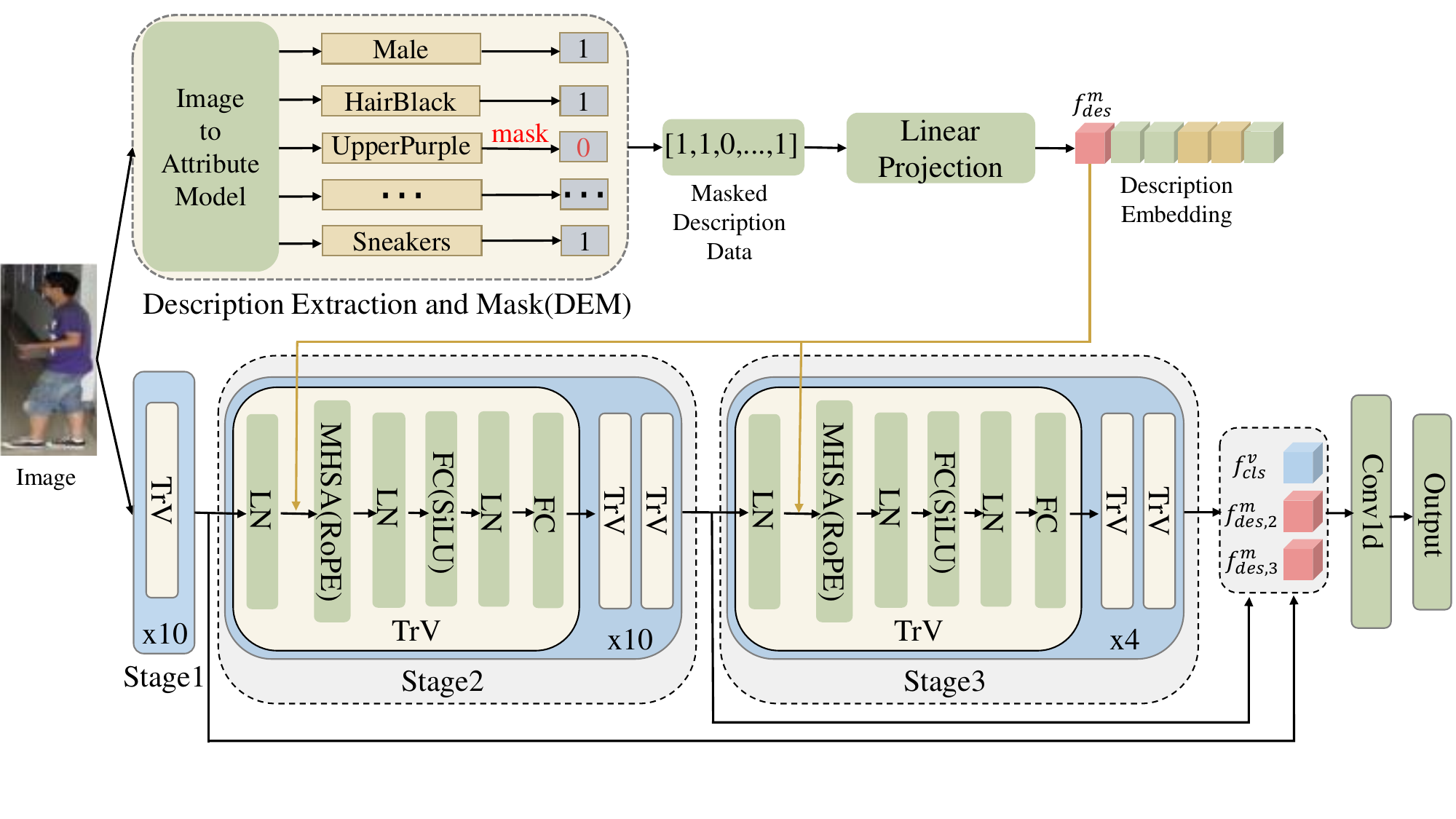}
	\caption{The framework of Masked Attribute Description Embedding (MADE) method. We first extract editable attribute description from the image through Description Extraction and Mask (DEM) module. After the cloth-related attribute descriptions are masked and converted into a binary vector, it is connected and embedded at different levels through Linear Projection to fuse with image features. Finally, we aggregate $f_{cls}^{v}$, $f_{des,2}^{m}$ and $f_{des,3}^{m}$ through Conv1D to obtain the person feature representation}
	\label{fig:method}
\end{figure*}

\subsection{Description Extraction and Mask (DEM)}
\label{3.2}

In CC-ReID, it is imperative for the model to disregard cloth information in the RGB image during input to learn cloth-insensitive features. DeSKPro~\cite{DeSKPro} and SAVS~\cite{SAVS} use the human parsing model to generate person parsing maps to remove clothes interference. However, processing the parsing map to obtain robust cloth-irrelevant information is more complicated. The description information of persons mainly describes the appearance and clothes of persons, so previous research rarely involves processing person description in CC-ReID. However, we could eliminate interference information from person's clothes in the model input by simply editing the attribute description. In MADE, for each input image $x_{i}$, we extract the description and operate the clothes mask through DEM, as shown in Fig. \ref{fig:method}.

DEM uses the extraction model to obtain personal descriptions suitable for cloth-changing datasets and performing mask processing. Specifically, we use SOLIDER~\cite{SOLIDER} trained on PETA\_ZS~\cite{petazs} to identify person attributes in cloth-changing datasets. SOLIDER~\cite{SOLIDER} is a human task visual pre-training model that adopts self-supervised training. We use it to obtain attribute descriptions of images. PETA\_ZS contains 19,000 images, including 8,705 individuals, each annotated with 61 binary and four multiclass attributes, 105 attribute labels in total. \textcolor{black}{They can be divided into categories such as gender, age, orientation, type of carried items, upper body color, upper body type, lower body color, lower body type, shoe color, and shoe type. We define upper body color, upper body type, lower body color, and lower body type as cloth-related attributes, while the others are defined as cloth-unrelated attributes.} Given a sample $x_{i}$, input SOLIDER to get personal attributes list, [$a_{1}^{i}$, $a_{2}^{i}$,...,$a_{N}^{i}$], and convert them into 0-1 binary vectors by position. We implement the cloth information mask operation by setting the clothes attributes to 0. And, the masked attribute description data [$m_{1}^{i}$, $m_{2}^{i}$,...,$m_{N}^{i}$] with the cloth information removed is obtained.

\textcolor{black}{Fig. \ref{fig:sam} presents some examples of pedestrian attributes extracted using SOLIDER \cite{SOLIDER}. Attribute recognition is a multi-classification task, where the attributes recognized for each pedestrian are not precisely the same. Among these attributes, those related to clothes (upper body color, upper body type, lower body color, and lower body type) are marked in black, while attributes unrelated to clothes are marked in blue. The cloth-related attributes are masked (set to 0) to obtain the final masked description data.}

In all experiments, we train the model with parameter settings that follow the original project~\cite{SOLIDER}.

\begin{figure*}[t]
	\centering
	\includegraphics[width=0.8\linewidth]{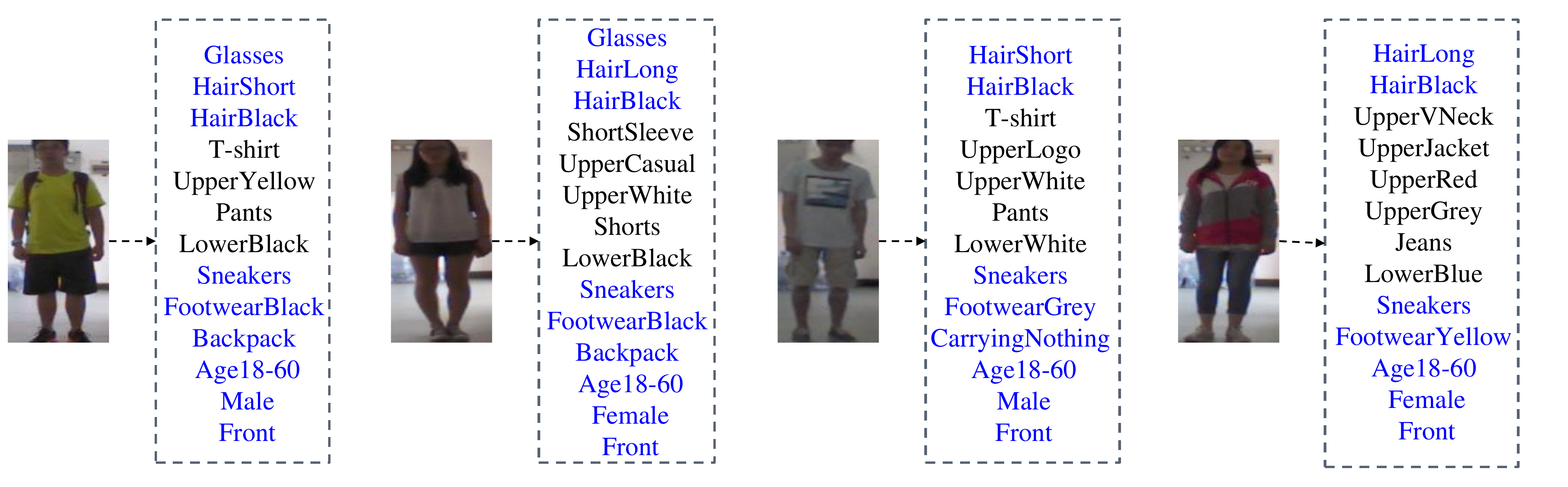}
	\caption{\textcolor{black}{Examples of pedestrian attribute lists extracted using SOLIDER (Attributes related to clothes are marked in black, while attributes unrelated to clothes are marked in blue). }
}
	\label{fig:sam}
\end{figure*}

\subsection{Description Embedding}
\label{3.3}

More than relying on appearance information is required to distinguish persons who change clothes accurately. Immutable multi-modal features can assist recognition~\cite{FSAM, CESD}. The advancement of Text-to-Image Person Retrieval demonstrates that additional text information can be fully leveraged when learning images of persons to enhance the final decision-making process. However, existing approaches often utilize a text encoder to handle entire sentences. We propose MADE to embed the masked attribute description vector directly into the Transformer block.

Given an image sample $x_{i}\in R^{H\times W\times C}$and its masked-description data $m_{i}$ = [$m_{1}^{i}$, $m_{2}^{i}$,\ldots,$m_{N}^{i}$], we integrate them into TrV Block~\cite{EVA02} to get the MADE framework, as shown in Fig. \ref{fig:method}. Considering the accuracy of the attribute extraction model, we introduce random 0-1 noise into the masked-description data (the discussion regarding the correctness of the attribute extraction model and the proportion of noise added is in section \ref{des}). TrV Block is an improved Vision Transformer structure. We use EVA02-large as the backbone, which has 24 layers of TrV Blocks and divides it into three stages (we discuss it in detail in section \ref{stnum}). First, we segment the image $x_{i}$ into a sequence of $N=H\times W/P^{2}$ fixed-size, where P represents the size of the patch, and then map the patch sequence through a trainable linear projection as one-dimensional notation \{${f_{j}^{v} |_{j=1}^{N}}$\}. After the injection of positional embedding and additional [CLS] token, the tokens sequence \{$f_{cls}^{v},f_{1}^{v},…,f_{N}^{v}$\} is input into the TrV block of $L$ layers of the first stage to model dependencies between each patch. Subsequently, $f_{cls}^{v}$ is extracted to represent the first stage's global image low-level feature representation. 

Then \textcolor{black}{masked-description data} $m_{i}$ is passed through Linear Projection \textcolor{black}{and expanded to three dimensions} to obtain the description feature \textcolor{black}{\{$f^{m}$\} aligning the third dimension with the same dimension as \{$f_{j}^{v} \mid_{j=1}^{N}$\}. After adding the extra token [DES], the attribute description sequence is \textcolor{black}{\{$f_{des}^{m}, f^{m}$\}}, as shown in Fig. \ref{fig:method}}. The description sequence is then connected to the image sequence, \textcolor{black}{\{$f_{j}^{vm} \mid_{j=1}^{N}$\} = \{$f_{des}^{m}, f^{m}, f_{1}^{v}, \ldots, f_{N}^{v}$\}}, and input into the second and third stages for training. In this process, image features and description features are embedded through connections and trained together to learn the relationship between images and attribute \textcolor{black}{descriptions} that mask cloth information. The interference of cloth-sensitive features can be removed through mask items and connection embedding, avoiding the problem of complex extraction and fusion of multi-modal biometric features.

In the model, we extract $f_{des}^{m}$ as a fusion representation of image and attribute description features. The class tokens output by the second and third stages, $f_{des,2}^{m}$ and $f_{des,3}^{m}$, respectively, represent the fusion of different levels of visual features and attribute description of clothes removal. We combine them with the output of $f_{cls}^{v}$ from the first stage to obtain the pedestrian feature representation of MADE through Conv1d aggregation.

\subsection{Loss Function and Inference}
\label{3.4}

In our experiments, we use cross-entropy loss without label smoothing and triplet loss. the loss function of MADE can be defined as follows:

\begin{equation}
	\mathcal{L}=\lambda_{1}\mathcal{L}_{id}+\lambda_{2}\mathcal{L}_{tri}
	\label{eq:loss}
\end{equation}
\textcolor{black}{where} $\mathcal{L}$ is the total loss function of the MADE method, $\mathcal{L}_{id}$ represents cross-entropy loss, and $\mathcal{L}_{tri}$ represents triplet loss. $\lambda_{1}$ and $\lambda_{2}$ are trade-off parameters used to balance each contribution. In our experiments, both $\lambda_{1}$ and $\lambda_{2}$ are empirically set to 1.0.

Cross-entropy loss $\mathcal{L}_{id}$ is defined as:

\begin{equation}
	\mathcal{L}_{id}=-\sum_{i=1}^N\log\frac{\exp(W_{y_i}x_{p}^i+b_{y_i})}{\sum_{k=1}^C\exp(W_kx_p^i+b_k)}
\end{equation}
where $N$ is the number of images in mini-batch, $y_{i}$ is the label of feature $x_{p}^{i}$ and $C$ is the number of classes.

Triplet loss function $\mathcal{L}_{tri}$ is defined as follows:

\begin{equation}
	\mathcal{L}_{tri}(I_{iA},I_{iP},I_{iN})=max\{0,M+D(I_{iA},I_{iP})-D(I_{iA},I_{iN})\}
\end{equation}
where $D(.)$ is the squared Euclidean distance in the embedding space, and $M$ is a parameter called the margin that adjusts the separation between pairs of distances: ($f_{iA},f_{iP}$) and ($f_{iA},f_{iN}$). $I_{iA}$, $I_{iP}$, and $I_{iN}$ are anchor images, positive samples, and negative sample images, respectively. The model learns to minimize the distance between more similar images and maximize the distance between dissimilar images.

For inference, for a given query $q_{i}$ and masked attribute description data $m_{i}$ = [$m_{1}^{i}$, $m_{2}^{i}$,...,$m_{N}^{i}$], we only use query images $q_{i}$ and discard $m_{i}$ to inference.

\section{Experiments}
\label{sec:exp}
\subsection{Datasets}

\textbf{PRCC}~\cite{Alpher01} is a dataset including person contour sketch proposed by Yang \emph{et al.}, including 221 persons and 33,698 images. The photos are taken by three cameras, A, B, and C, respectively, and the clothes of persons in cameras A and B do not change. Camera C takes pictures at different times, and the clothes of persons are different from those in cameras A and B. There are about 50 images of each person under each camera view.

\textbf{LTCC}~\cite{CESD} is a dataset captured by 12 cameras for two months, including 162 persons and 15,138 images. The dataset is divided into two subsets: persons with changing clothes, including 91 people, 415 sets of different clothes, and 14,756 images; the set with consistent clothes, including 61 people and 2,382 images.

\textbf{Celeb-reID-light}~\cite{Celeb-reID-light}  contains 290 persons and 10,842 images. This dataset comes from the snapshots of celebrities on the Internet. Everyone in the dataset has about 20 pictures of different clothes, and people do not wear the same clothes.

\textbf{LaST}~\cite{LaST} is a large-scale dataset from over 2,000 movies in 8 countries. It includes 10,862 persons and 228,166 images. The training set has 5,000 identities and 71,248 images, the validation set has 56 identities and 21,379 images, and the test set has 5,806 identities and 135,529 images.

For the cloth-changing datasets PRCC~\cite{Alpher01} and LTCC~\cite{CESD}, we follow their respective evaluation protocols and evaluate the performance under the cloth-changing and standard settings. \textcolor{black}{In Figure \ref{fig:dataset}, we present examples of the datasets this paper used.}

\begin{figure*}[t]
	\centering
	\includegraphics[width=0.8\linewidth]{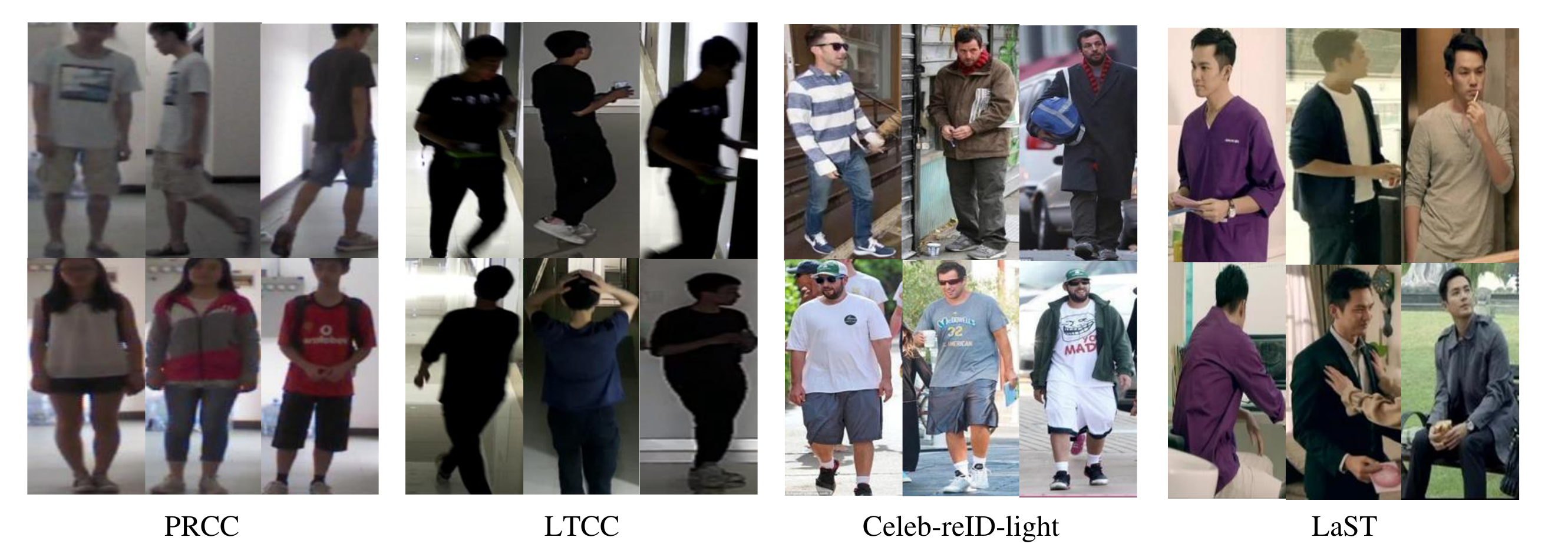}
	\caption{\textcolor{black}{Examples of the datasets this paper used. }
}
	\label{fig:dataset}
\end{figure*}

We adopt the standard metrics used in most of the person re-identification literature, namely the cumulative matching curve (CMC), to generate ranking accuracy and the mean average precision (mAP). We report rank-1 accuracy and mean average precision (mAP) on all datasets for evaluation.

\subsection{Person Attribute Analysis}

In order to explore whether the irrelevant attributes of clothes are retained when persons change clothes and are captured across cameras, and the proportion of retention. In the experiment, we evaluated the retention ratios of attributes in the training and test sets for the four datasets: PRCC, LTCC, Celeb-ReID-light, and LaST.

According to Fig. \ref{fig:par}, we can observe that for PRCC and LTCC, which are both collected from real-world scenarios with short data collection periods and fixed scopes, the biological attributes of pedestrians are maintained at a relatively high proportion. Due to the challenging style of the LTCC for attribute recognition models, in the ablation experiments of LTCC (section \ref{des}), the addition of the DEM module only marginally improves recognition accuracy. We discuss the impact of attribute recognition model accuracy on the experimental results in section \ref{des}. Surprisingly, for Celeb-ReID-light and LaST, which originate from internet images, their biological attribute features are also maintained at a high proportion. In Fig. \ref{fig:par}, we compute the average retention ratio for the four datasets. It can be observed that biological attributes such as age, gender, and hairstyle usually remain unchanged in the short term across different cameras. Additionally, even if individuals change clothes, features such as shoe type and color typically remain stable. These findings suggest that these attributes may be crucial for models to learn cloth-agnostic features, indicating that leveraging these stable attributes could aid models in better understanding and identifying individuals regardless of clothes variations.

\begin{figure*}[t]
\begin{center}
    \centering
    \subfigure[]{
        \begin{minipage}[b]{0.19\linewidth}
            \centering
            \includegraphics[width=\linewidth]{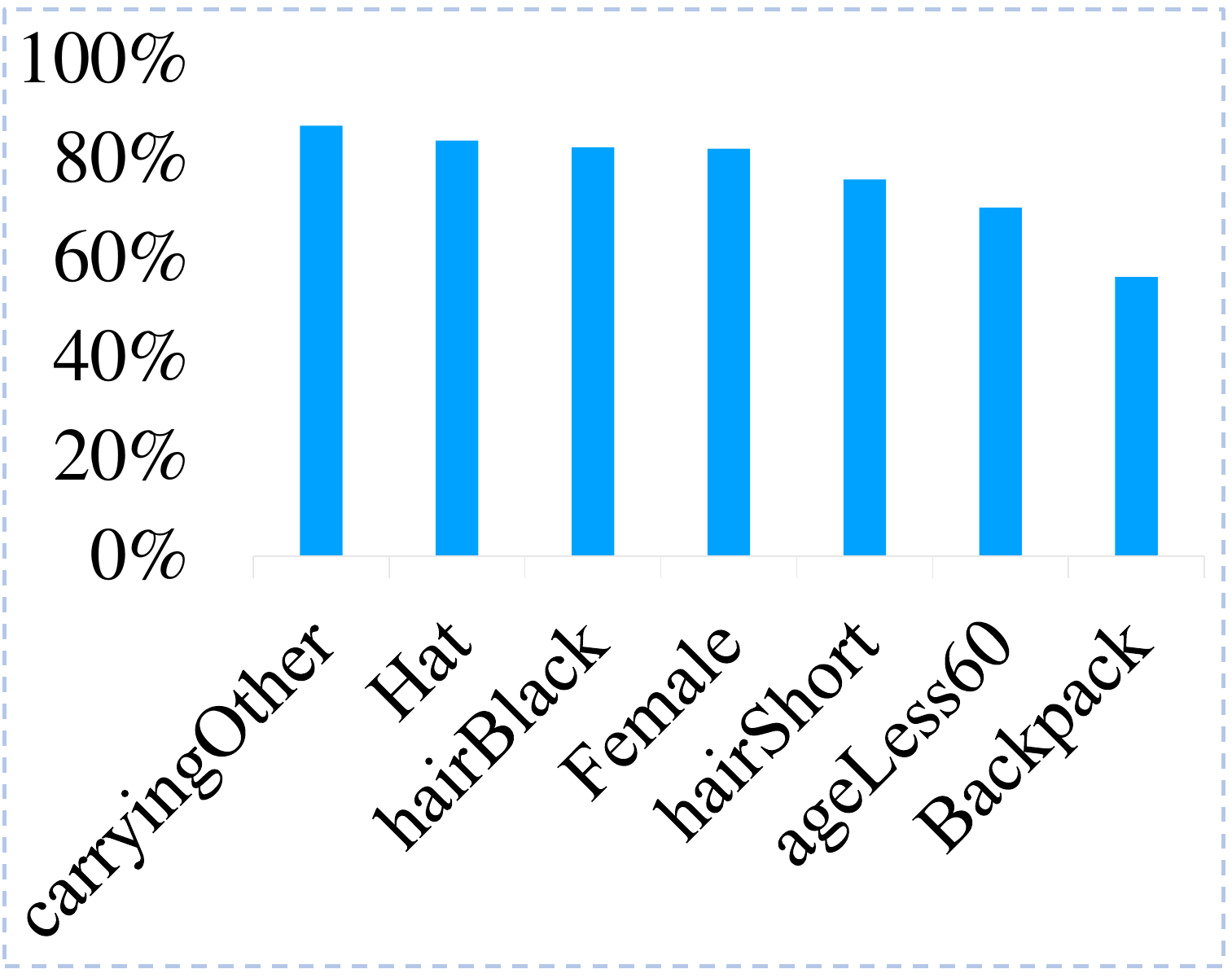}  
        \end{minipage}
    }%
    \subfigure[]{
        \begin{minipage}[b]{0.19\linewidth}
            \centering
            \includegraphics[width=\linewidth]{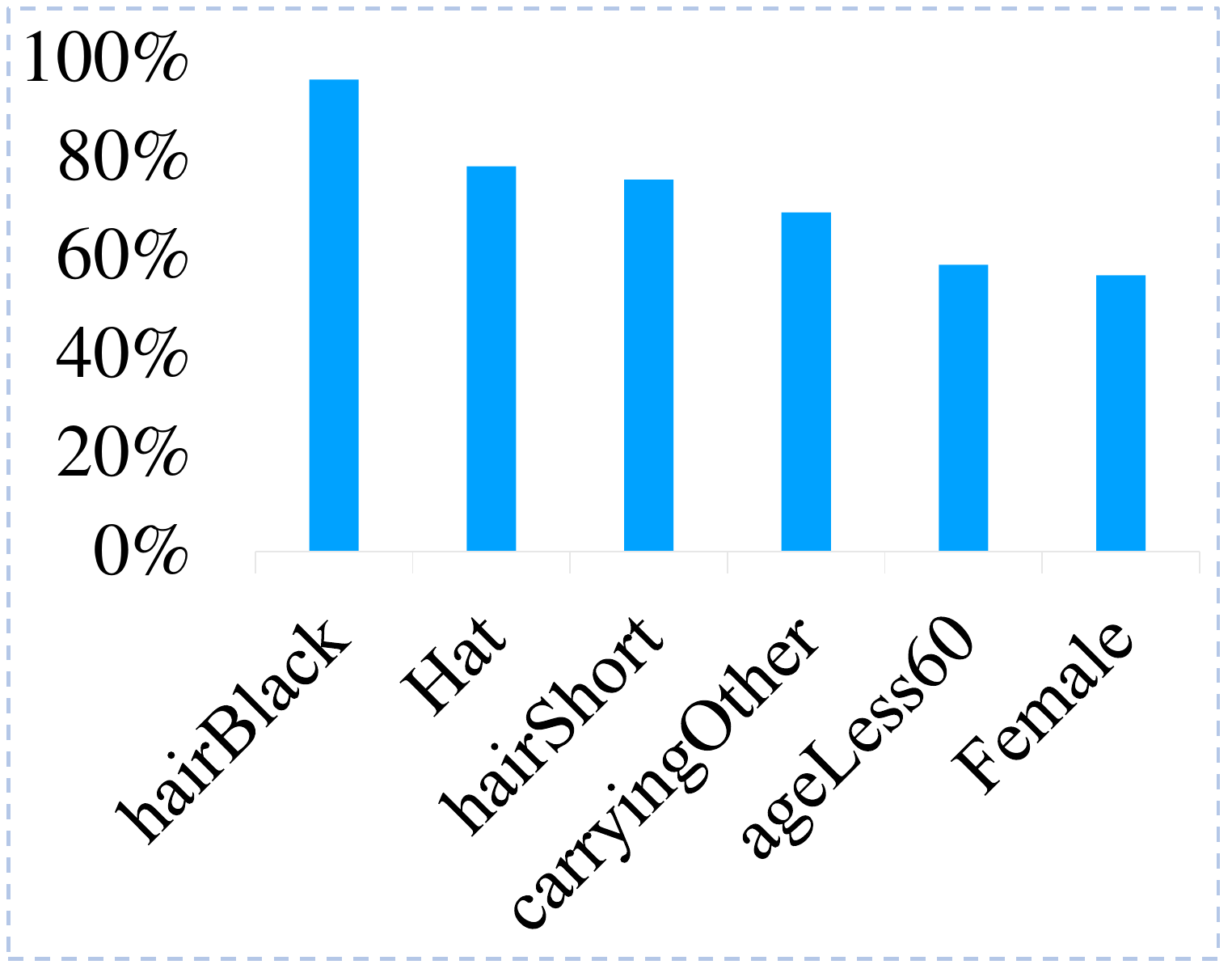}  
        \end{minipage}
    }%
    \subfigure[]{
        \begin{minipage}[b]{0.19\linewidth}
            \centering
            \includegraphics[width=\linewidth]{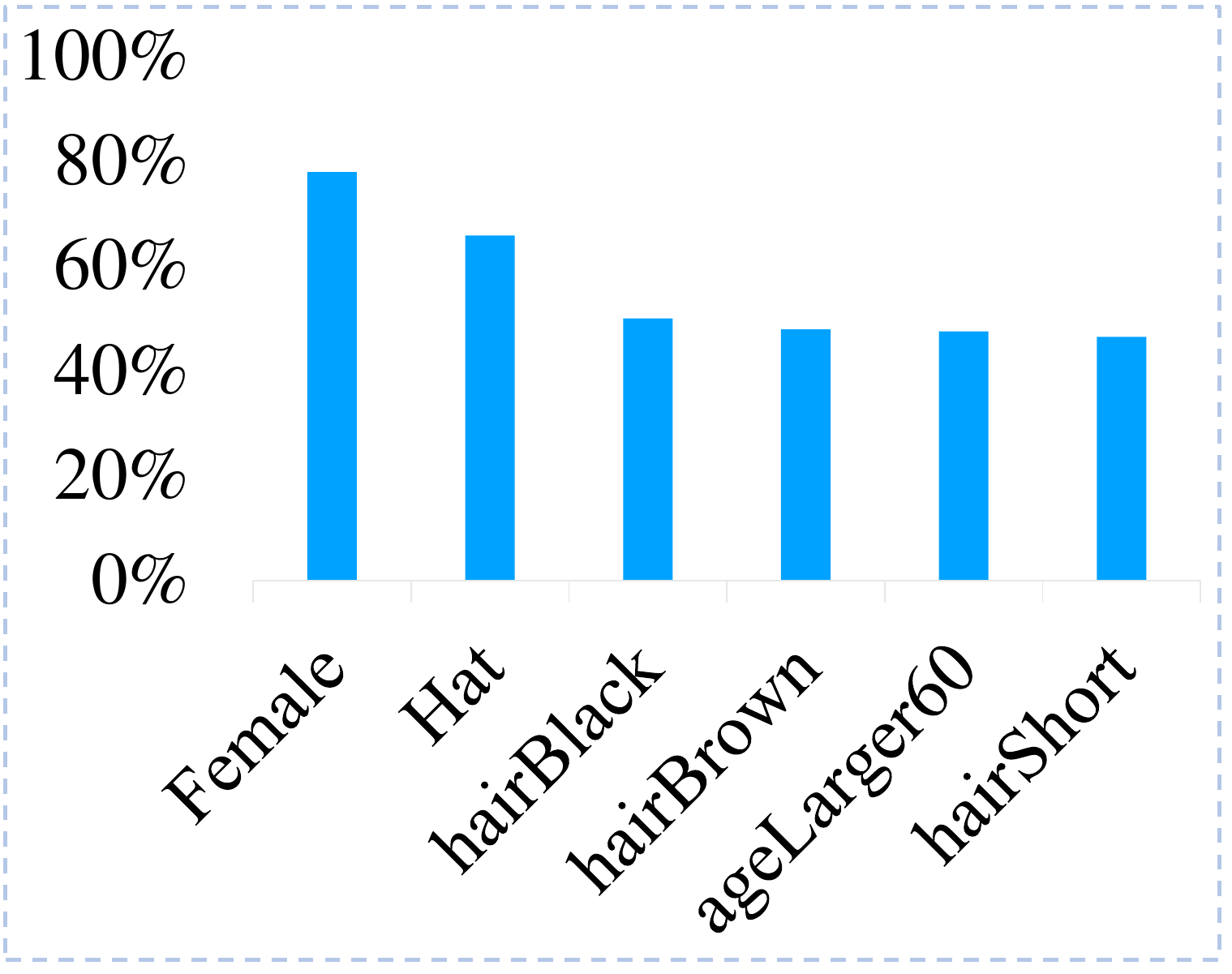}  
        \end{minipage}
    }%
    \subfigure[]{
        \begin{minipage}[b]{0.19\linewidth}
            \centering
            \includegraphics[width=\linewidth]{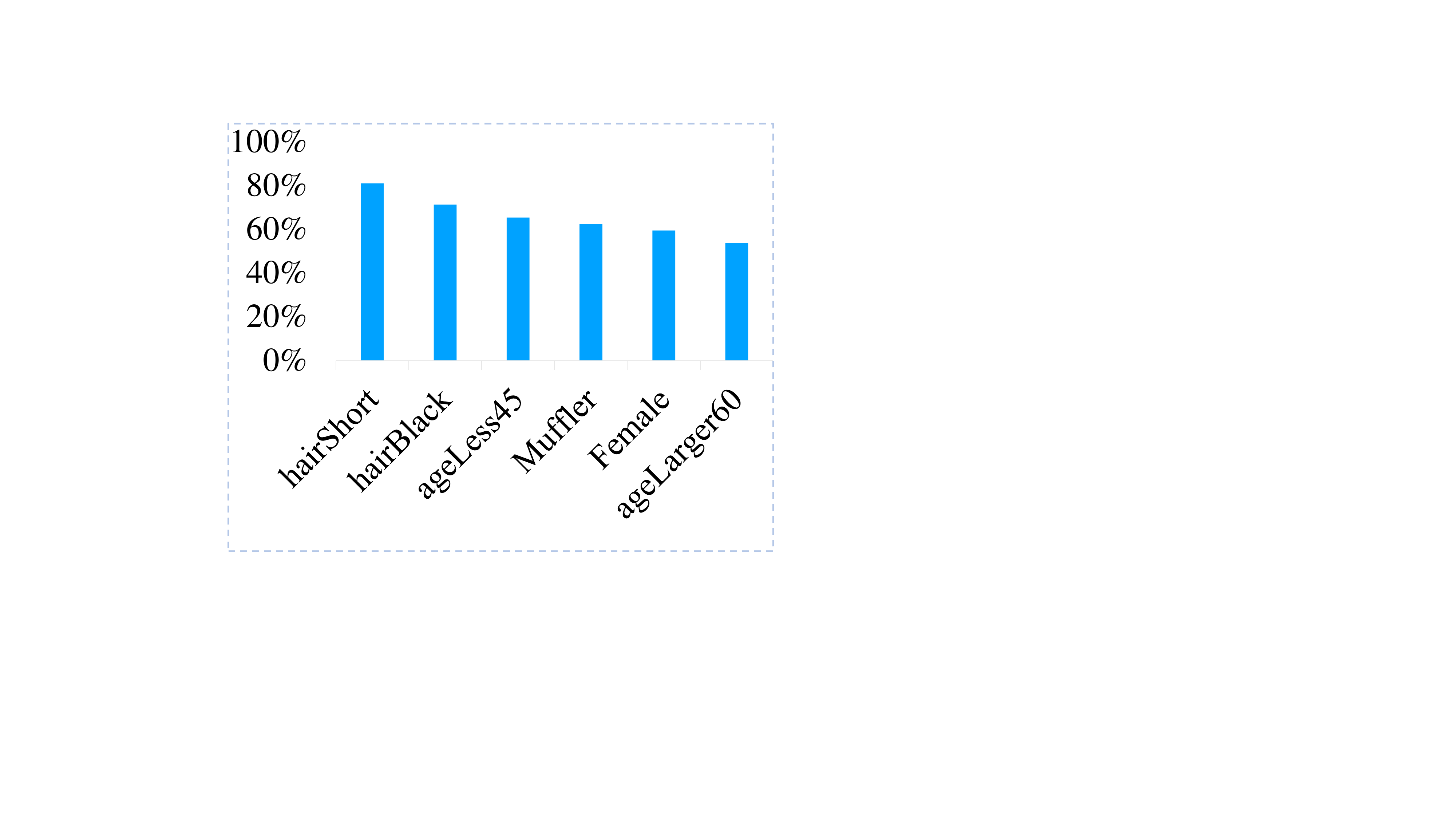}  
        \end{minipage}
    }%
    \subfigure[]{
        \begin{minipage}[b]{0.19\linewidth}
            \centering
            \includegraphics[width=\linewidth]{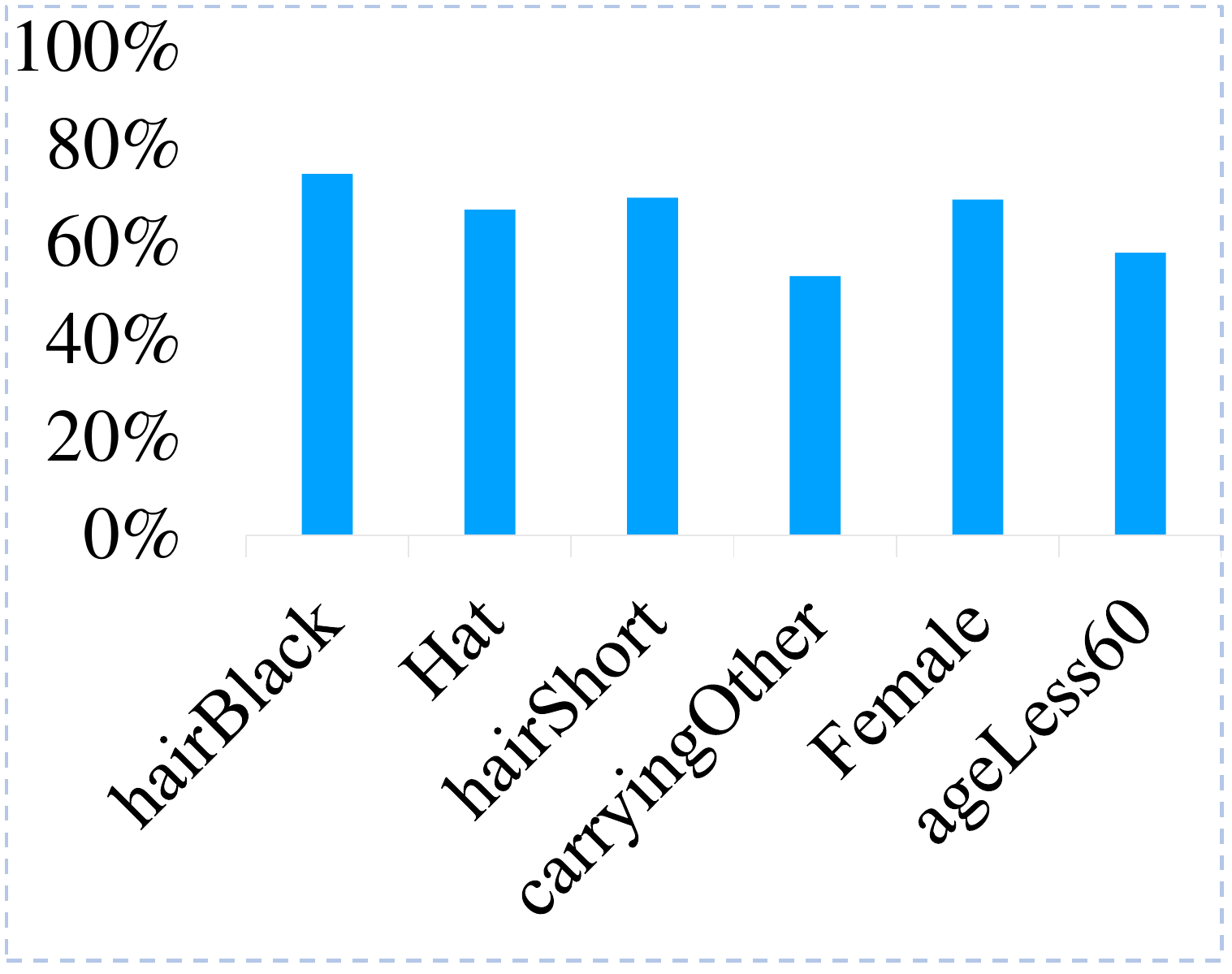}  
        \end{minipage}
    }%
    \end{center}
    \caption{Retain ratio of clothes irrelevant person attributes in each dataset. (a) PRCC, (b) LTCC, (c) Celeb-reID-light, (d) LaST, and (e) Average statistic.}
    \label{fig:par}
\end{figure*}

\subsection{Implementation Details}
\label{Implementation Details}
The input images are resized to 224$\times$224 for all datasets. We divide the 24 layers of EVA02-large~\cite{EVA02} into three stages, and the number of layers of the Trv blocks is 10, 10, and 4, respectively(We discussed in section \ref{stnum}). We use LayerNorm~\cite{LayerNorm} to normalize features. For data augmentation, we employ random cropping and random erasing ~\cite{56}. Due to the limit of GPU memory, the batch size is set to 8, each batch includes two different people, and the number of images for each person is 4. The SGD optimizer is employed in the optimization process, and 60 epochs are required. Moreover, the weight decay for the experiment is $5e^{-2}$. The warmup learning rate is initially set to $7.8125e^{-7}$. The learning rate is initially set to $2e^{-5}$ and divided by 100 at 40 and 60 epochs. The optimal parameter values are directly used for the other datasets without tuning.

\subsection{Experimental Comparison}

For RRCC and LTCC, we combine our proposed MADE method with some cloth-changing re-identification methods (i.e., SPT+ASE~\cite{Alpher01}, GI-ReID~\cite{GI-ReID}, CESD~\cite{CESD}, RCSANet~\cite{RCSANet}, 3DSL~\cite{3DSL}, FSAM~\cite{FSAM}, BSGA+CRE~\cite{Muetal}, CAL~\cite{CAL}, DCR-ReID~\cite{DCR-ReID}, CCFA~\cite{CCFA}, AIM~\cite{AIM} and chan \emph{et al.}~\cite{ACMMM23}) were compared. We compare Celeb-reID-light with four cloth-changing re-identification methods (RCSANet~\cite{RCSANet}, MBUNet~\cite{MBUNet}, IRANet~\cite{IRANet}, and DeSKPro~\cite{DeSKPro}) and some traditional methods. We compare LaST with CAL and some traditional methods. It is worth noting that among these CC-ReID methods, SPT+ASE, GI-ReID, CESD, 3DSL, FSAM, BSGA+CRE, and DCR-ReID all integrate person multi-modal biometric features into the model to remove clothes interference. CAL and AIM mine the information of original RGB images. CCFA adopts the feature enhancement method. The method proposed by Chen \emph{et al.}. is based on the GAN network. Considering the accuracy of the attribute extraction model, our experimental results are obtained under the premise of introducing 10\% random 0-1 noise into the masked-attribute description. Discussions on the accuracy of the attribute extraction model and the proportion of noise can be found in Section \ref{des}.

\begin{table*}
	\caption{{Evaluations on the PRCC and LTCC datasets (\%), where "sketch," "pose," "sil.", "parsing" and "3D" denote contour sketches, keypoints, silhouettes, human parsing, and 3D shape information. Bold and underlined numbers are the top two scores.}}
	\label{tab:table1}
	\normalsize
	\begin{center}
		\begin{adjustbox}{width=0.8\linewidth}
			\begin{tabular}{c| c| c|cc|cc|cc|cc}
				\hline
				\multirow{3}{*}{Method} & \multirow{3}{*}{Venue} & \multirow{3}{*}{Modality}  & \multicolumn{4}{c|}{PRCC} & \multicolumn{4}{c}{LTCC}\\
				\cline{4-7} \cline{8-11}
				&  &  & \multicolumn{2}{c|}{CC} & \multicolumn{2}{c|}{SC} & \multicolumn{2}{c|}{CC} &\multicolumn{2}{c}{Genral}\\
				\cline{4-11}
				&  &  & rank-1 & mAP & rank-1 & mAP & rank-1 & mAP & rank-1 & mAP \\
				\hline
				SPT+ASE ~\cite{Alpher01} & TPAMI 19 & Sketch & 34.4 & - & 64.2 & - & - & - & - & - \\
				CESD ~\cite{CESD} & ACCV 20 & RGB+pose & - & - & - & - & 26.1 & 12.4 & 71.4 & 34.3 \\
				RCSANet ~\cite{RCSANet} & ICCV 21 & RGB & 48.6 & 50.2 & \textbf{100.0} & 97.2 & - & - & - & - \\
				3DSL ~\cite{3DSL} & CVPR 21 & RGB+pose+sil.+3D & - & 51.3 & - & - & 31.2 & 14.8 & - & - \\
				FSAM ~\cite{FSAM} & CVPR 21 & RGB+pose+sil. & 54.5 & - & 98.8 & - & 38.5 & 16.2 & 73.2 & 35.4 \\
				GI-ReID ~\cite{GI-ReID} & CVPR 22 & RGB+sil. & - & 37.5 & - & - & 23.7 & 10.4 & 63.2 & 29.4 \\
				BSGA+CRE ~\cite{Muetal} & BMVC 22 & RGB+parsing & \underline{61.8} & \underline{58.7} & \underline{99.6} & 97.3 & - & - & - & - \\
				CAL ~\cite{CAL} & CVPR 22 & RGB & 55.2 & 55.8 & \textbf{100.0} & \underline{99.8} & 40.1 & 18.0 & 74.2 & 40.8 \\
				CCFA ~\cite{CCFA} & CVPR 23 & RGB & 61.2 & 58.4 & \underline{99.6} & 98.7 & \underline{45.3} & \underline{22.1} & 75.8 & \underline{42.5} \\
				AIM ~\cite{AIM} & CVPR 23 & RGB & 57.9 & 58.3 & \textbf{100.0} & \textbf{99.9} & 40.6 & 19.1 & \underline{76.3} & 41.1 \\
				DCR-ReID ~\cite{DCR-ReID} & TCSVT 23 & RGB+parsing & 57.2 & 57.4 & \textbf{100.0} & 99.7 & 41.1 & 20.4 & 76.1 & 42.3 \\
				chan \emph{et al.} ~\cite{ACMMM23} & ACM 23 & RGB & 58.4 & 58.6 & \textbf{100.0} & 99.7 & 32.9 & 15.4 & 73.4 & 36.9 \\ 
				\hline
				MADE &  & RGB+description & \textbf{64.3} & \textbf{59.1} & \textbf{100.0} & 98.6 & \textbf{47.4} & \textbf{24.4} & \textbf{84.2} & \textbf{48.2 }\\
				
			\end{tabular}
		\end{adjustbox}
	\end{center}
\end{table*}

\textbf{Results on PRCC.} We compare our method with twelve cloth-changing re-identification methods on PRCC in Table \ref{tab:table1}. We can notice that our method outperforms all other methods in the cloth-changing setting. Compared with AIM, the method of mining original RGB images in the cloth-changing setting, the rank-1 of our method increased by 6.4\%, and the mAP increased by 0.8\%. These shows that using multi-modal attribute description in CC-ReID to assist re-identification effectively improves results. Compared with the best method using multi-modal biometrics, BSGA+CRE, in the cloth-changing setting, the rank-1 increased by 2.5\%, and the mAP increased by 0.4\%. It shows that our method uses editable multi-modal attribute information and has better results when it is more convenient to remove cloth information than biological information. Data augmentation can significantly improve the model improvement effect, and our method is even better than CCFA, which uses feature enhancement. In the cloth-changing setting, the rank-1 increased by 3.1\%, and the mAP increased by 0.7\%.

\textbf{Results on LTCC.} We compare our method with eight cloth-changing re-identification methods on LTCC in Table \ref{tab:table1}. Our method outperforms all other methods in both cloth-changing and general settings. Compared with CCFA, in the cloth-changing setting, The rank-1 increased by 2.1\%, and the mAP increased by 2.3\%. In the general setting, The rank-1 increased by 8.4\%, and the mAP increased by 5.7\%. Compared with AIM, in the cloth-changing setting, The rank-1 increased by 6.8\%, and the mAP increased by 5.3\%. In the general setting, The rank-1 increased by 7.9\%, and the mAP increased by 7.1\%. Compared with chan \emph{et al.}, in the cloth-changing setting, The rank-1 increased by 14.5\%, and the mAP increased by 9.0\%. In the general setting, The rank-1 increased by 10.8\%, and the mAP increased by 11.3\%. It shows that our method embedding fused image features with attribute descriptions is better than using GAN networks to mine cloth-irrelated features of original images.

\begin{table}
	\caption{{Evaluations on Celeb-reID-light(\%). Bold and underlined numbers are the top two scores.}}
	\label{tab:tablec}
	\normalsize
	\begin{center}
		\begin{adjustbox}{width=0.8\linewidth}
			\begin{tabular}{c|c|c|cc}
				\hline	 
				\multirow{3}{*}[1ex]{Method Type} & \multirow{3}{*}[1ex]{Method} & \multirow{3}{*}[1ex]{Venue} & \multicolumn{2}{c}{Celeb-reID-light} \\
				\cline{4-5} 
				& & & rank-1 & mAP  \\
				\hline 
				\multirow{5}{*}{Traditional}
				& OSNet ~\cite{cc101} &ICCV 19 & 21.3 & 11.7 \\
				& DG-Net ~\cite{tr43}& CVPR 19 & 23.5 & 12.6 \\
				& BoT(resnet50) ~\cite{cc110}  & CVPRW 19 & 24.2  & 13.6 \\
				& AGW(resnet50{\_}nl) ~\cite{su66} & TPAMI 21 & 30.2 & 15.4 \\
				& TransReID ~\cite{cc112}& ICCV 21 & 31.3  & 18.6 \\
				\hline 
				\multirow{6}{*}{CC-ReID} 
				& RCSANet ~\cite{RCSANet} & CVPR 21  & 29.5 & 16.7 \\
				& MBUNet ~\cite{MBUNet} & ICME 22  & 33.9 & 21.3\\
				& IRANet ~\cite{IRANet} & IVC 22  &46.2 & 25.4 \\
				& DeSKPro ~\cite{DeSKPro} & ICIP 22 & \underline {52.0} & \underline {29.8}  \\
				\cline{2-5} 
				& MADE &  & \textbf{72.0} & \textbf{52.3}  \\
			\end{tabular}
		\end{adjustbox}
	\end{center}
\end{table}

\begin{table}
	\caption{{Evaluations on LaST(\%). Bold and underlined numbers are the top two scores.}}
	\label{tab:tablel}
	\normalsize
	\begin{center}
		\begin{adjustbox}{width=0.8\linewidth}
			\begin{tabular}{c|c|c|cc}
				\hline	 
				\multirow{3}{*}[1ex]{Method Type} & \multirow{3}{*}[1ex]{Method} & \multirow{3}{*}[1ex]{Venue} & \multicolumn{2}{c}{LaST} \\
				\cline{4-5} 
				& & & rank-1 & mAP  \\
				\hline 
				\multirow{5}{*}{Traditional}
				& OSNet ~\cite{cc101} & ICCV 19 & 64.3  & 21.0 \\  
				& BoT ~\cite{cc110} & CVPRW 19 & 67.1  & 23.6 \\ 
				& HOReID ~\cite{cc125} & CVPR 20 & 68.3  & 25.5 \\ 
				& Top-DB-Net ~\cite{cc124} & ICPR 20 & 69.4  & 25.0 \\ 
				& CtF ~\cite{cc126} & ECCV 20 & 70.0  & 26.5 \\
				\hline 
				\multirow{3}{*}{CC-ReID} 
				& CAL ~\cite{CAL} & CVPR 22 & \underline{73.7} & \underline{28.8} \\
				\cline{2-5}
				& MADE &  & \textbf{79.0} &  \textbf{40.9} \\
			\end{tabular}
		\end{adjustbox}
	\end{center}
\end{table}

\textbf{Results on Celeb-reID-light and LaST.} We compare our method with some traditional and cloth-changing re-identification methods on the two datasets in Table \ref{tab:tablec} and Table \ref{tab:tablel}. Our method outperforms all previous methods. The face is the most direct information for re-identification. For Celeb-reID-light, our method is even better than DeSKPro, which uses facial features. The rank-1 increased by 20.0\%, and the mAP increased by 22.5\%. For LaST, our method is better than CAL. The rank-1 increased by 5.3\%, and the mAP increased by 12.1\%. LaST is a large and challenging dataset that requires high model training complexity. Currently, there are few CC-ReID methods tested using LaST. It shows that our method is effective, has low model complexity, and can achieve testing of large datasets. 

\subsection{Ablation Study}

We use EVA02-large ~\cite{EVA02} as the baseline and also use the loss function in section \ref{3.4} for supervision. In this section, we explore the role of masked attribute description on the model's learning of clothes-irrelevant features and the impact of different layer numbers in EVA02-large.

\subsubsection{\textcolor{black}{Effectiveness of Mask Attribute and Influence of Attribute Detection Accuracy}}
\label{des}
\textcolor{black}{We embedded the attribute description vector after masking cloth information into the baseline. Considering the accuracy of the attribute extraction model, we also introduced a certain proportion of random 0-1 noise into the masked attribute and summarized the experimental results in Table \ref{tab:table3}.} To verify the effectiveness of attribute description in improving re-identification results. We conducted experiments on the cloth-changing setting of PRCC, LTCC and Celeb-reID-light. In all datasets, the performance of adding masked attribute descriptions is almost higher than the baseline.

The results of adding masked-attribute descriptions were higher than the baseline. After masking the cloth information in the attribute description \textcolor{black}{without noise} and embedding, for PRCC, the Rank-1 increased by 4.2\%, and the mAP increased by 5.2\%. For LTCC, the Rank-1 increased by 2.3\%, and the mAP increased by 1.5\%. \textcolor{black}{Considering the accuracy issues of the attribute extraction model, we introduced random noise of 5\%, 10\%, 15\%, and 20\% separately into the attribute descriptions of every person in the PRCC, LTCC, and Celeb-reID-light datasets. Due to the varying effects of different noise levels on the improvement of attribute description data across different datasets, we report the average results of the experiments. When the noise level was 10\%, the average rank-1 value across the three datasets was the highest, so we selected the experimental results with 10\% noise as the final result. Specifically, compared to the baseline, when 10\% noise was introduced, the rank-1 for PRCC increased by 2.3\%, for LTCC increased by 3.8\%, and for Celeb-reID-light increased by 4.2\%.} The model is more accurate in person re-identification, indicating that this method can compel the model to discard cloth information and learn cloth-insensitive features.

Then we discuss the impact of the accuracy of attribute detection models on experimental results. The attribute detection model we used is SOLIDER~\cite{SOLIDER}. \textcolor{black}{It has achieved excellent performance on widely used pedestrian attribute recognition datasets PETA\_ZS~\cite{petazs}, RAP\_ZS~\cite{petazs}, and PA100K~\cite{pa100k}, with mean accuracy (mA) of 76.4, 76.4, and 86.4, respectively~\cite{SOLIDER}. Although its accuracy on the attribute recognition dataset did not achieve complete correctness, our method is robust to a certain proportion of errors in attribute detection. Since the cloth-changing dataset lacks attribute labels, it is not feasible to directly measure the accuracy of attribute recognition. Hence, we introduce random noise of 5\%, 10\%, 15\%, and 20\% separately into the attribute description of every person in PRCC, LTCC, and Celeb-reID-light. The aim is to investigate the influence on person re-identification results when the attribute recognition model is not sufficiently accurate. Following the introduction of noise, the experimental results of PRCC and LTCC under the cloth-changing setting and Celeb-reID-light are shown in Table \ref{tab:table3}.}

\textcolor{black}{In the LTCC dataset, appropriately adding random noise can improve re-identification accuracy, but adding more than 10\% noise leads to a slight deterioration in results. However, in the PRCC dataset, adding a certain proportion of noise generally leads to a slight decrease in re-identification results, while adding 20\% noise increases the rank-1 by 5.7\% compared to the baseline. One possible reason is the difference in dataset styles. As shown in Fig. \ref{fig:samstyle}, pedestrian images in LTCC are generally darker overall, resulting in lower attribute recognition accuracy. Adding appropriate random noise can enhance pedestrian attribute recognition performance. On the other hand, images in PRCC have more apparent colors and pedestrian attributes are relatively easier to identify than those in LTCC. Therefore, adding a certain proportion of random noise decreases recognition performance. SOLIDER training uses PETA\_ZS, a dataset collected from real-world scenarios, as shown in Fig. \ref{fig:samstyle}(a), whereas Celeb-reID-light is collected from the internet, as shown in Fig. \ref{fig:samstyle}(d). The difference in dataset styles may lead to higher re-identification performance when noise is added to Celeb-reID-light compared to when no noise is added.}

\textcolor{black}{In addition, Table \ref{tab:table_scale} compares our model with the baseline on the PRCC dataset in terms of experimental scale. The experiment follows the setup described in Section \ref{Implementation Details}. We calculated the time required for the model to train one epoch (batch size = 8), the FLOPs for a single image input to the model, and the model's parameters. During the testing phase, we removed the attribute description information and calculated the time required for testing with a batch size of 128.}

\begin{table*}
	\caption{{\color{black} Ablation studies of attribute description of MADE in cloth-changing setting on PRCC, LTCC and Celeb-reID-light. Where m-att means masked attribute}}
	\label{tab:table3}
	\normalsize
	\begin{center}
		\begin{adjustbox}{width=0.7\linewidth}
			\begin{tabular}{c|cc|cc|cc|cc}
				\hline
				\multirow{3}{*}[1ex]{\color{black} Method}  & \multicolumn{2}{c|}{\color{black} PRCC} & \multicolumn{2}{c|}{\color{black} LTCC} & \multicolumn{2}{c|}{\color{black} Celeb-reID-light} & \multicolumn{2}{c}{\color{black} Average} \\
				\cline{2-9} 
				& \color{black} rank-1 & \color{black} mAP & \color{black} rank-1 & \color{black} mAP & \color{black} rank-1 & \color{black} mAP & \color{black} rank-1 & \color{black} mAP \\
				\hline
				\color{black} baseline & \color{black} 62.0 & \color{black} 57.8 & \color{black} 43.6 & \color{black} 23.5 & \color{black} 65.4& \color{black}46.8 & \color{black} 57.0 & \color{black} 42.7\\
				\hline
				\color{black} baseline w/ m-att (0\% noise) & \color{black} 66.2 & \color{black} 63.0 & \color{black} 45.9 & \color{black}  25.0 & \color{black} 67.3 & \color{black} 47.3
    & \color{black} 59.8 & \color{black}45.1\\
             \color{black} baseline w/ m-att (5\% noise) & \color{black} 65.7 & \color{black} 61.5 & \color{black} 46.4 & \color{black}  23.3 & \color{black} 70.9 & \color{black} 51.1
             & \color{black} 61.0 & \color{black}45.3\\\
             \color{black} baseline w/ m-att (10\% noise) & \textcolor{black}{64.3} & \textcolor{black}{59.1} & \textcolor{black}{47.4} &  \textcolor{black}{24.4} & \color{black} 72.0 & \color{black} 52.3
              & \color{black} 61.2 & \color{black}45.3\\
             \color{black} baseline w/ m-att (15\% noise) & \textcolor{black}{65.6} & \textcolor{black}{60.9} & \textcolor{black}{42.9} &  \textcolor{black}{21.1} & \color{black} 69.6 & \color{black} 48.2
             & \color{black} 59.4 & \color{black}43.4\\
             \color{black} baseline w/ m-att (20\% noise) & \textcolor{black}{67.7} & \textcolor{black}{64.7} & \textcolor{black}{40.1} &  \textcolor{black}{19.0} & \color{black} 67.4 & \color{black} 47.3 
             & \color{black} 58.4 & \color{black} 43.7 \\
			\end{tabular}
		\end{adjustbox}
	\end{center}
\end{table*}

\begin{figure*}[t]
	\centering
	\includegraphics[width=0.8\linewidth]{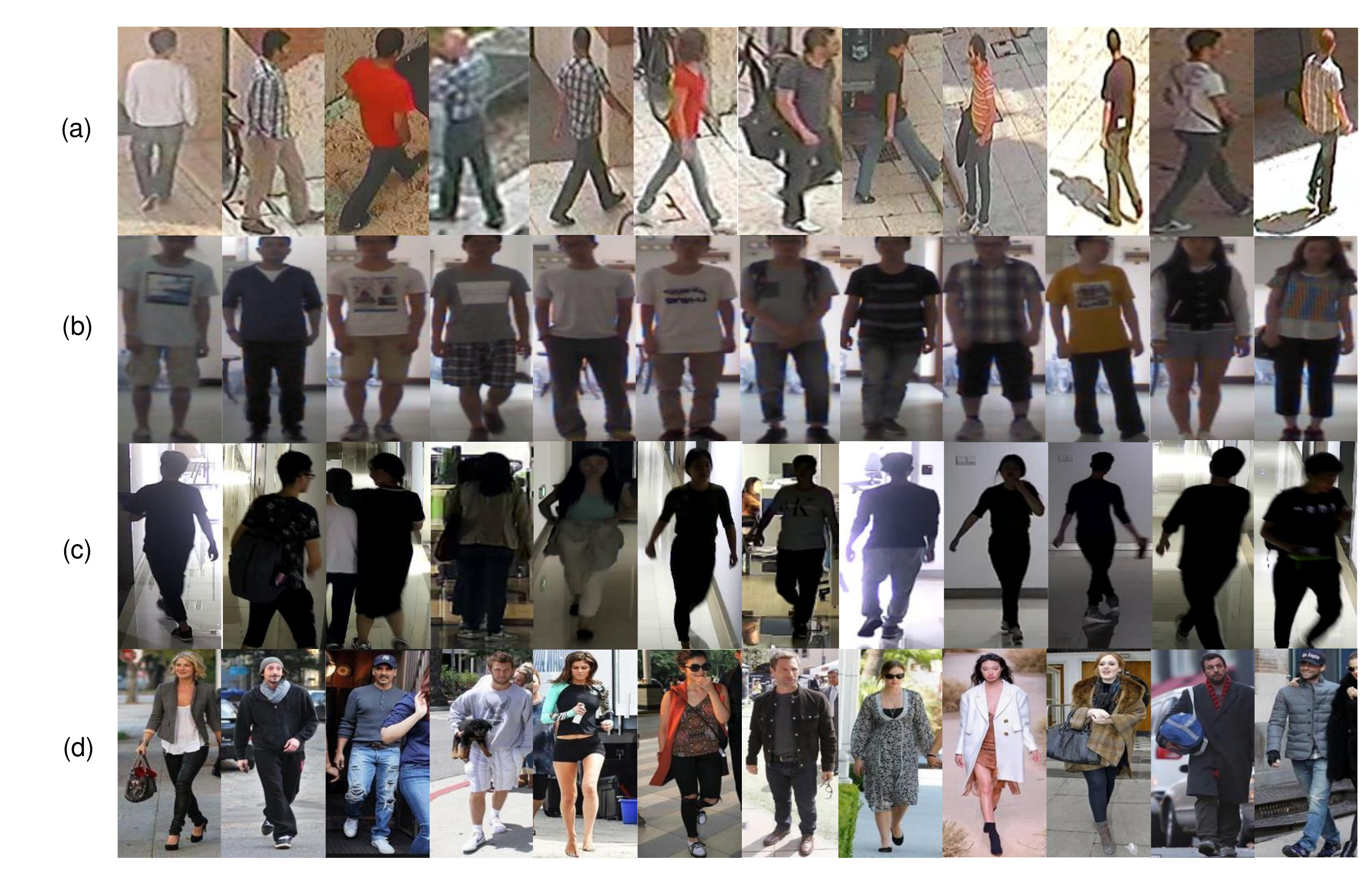}
	\caption{\textcolor{black}{(a) Examples of PETA\_ZS. (b) Examples of PRCC. (c) Examples of LTCC. (d) Examples of Celeb-reID-light. PETA\_ZS, PRCC, and LTCC are datasets collected from real-world scenarios.} Photos of LTCC exhibit an overall dark style, which may adversely affect the accuracy of attribute recognition models. PRCC dataset features a bright style that facilitates the identification of pedestrian attributes. \textcolor{black}{Celeb-reID-light is a dataset collected from the internet.}
}
	\label{fig:samstyle}
\end{figure*}

\begin{table}
	\caption{{\color{black}The comparison of the experimental scale between the Baseline and our model on the PRCC dataset.}}
	\label{tab:table_scale}
	\normalsize
	\begin{center}
		\begin{adjustbox}{width=\linewidth}
			\begin{tabular}{c|ccc|ccc}
				\hline
				\multirow{3}{*}[1ex]{\color{black}Method}  & \multicolumn{3}{c|}{\color{black}Training} & \multicolumn{3}{c}{\color{black}Testing}   \\
				\cline{2-7} 
				& {\color{black}Epoch time} & {\color{black}FLOPs} & {\color{black}Params} & {\color{black}CPU time} & {\color{black}FLOPs} & {\color{black}Params}  \\
				\hline
				{\color{black}baseline} & {\color{black}7m42s} & {\color{black}162.4GFlops} & {\color{black}304.1M} & \textcolor{black}{6h0m} & {\color{black}162.4GFlops} & {\color{black}304.1M} \\
				{\color{black}ours} & {\color{black}10m20s} & {\color{black}162.8GFlops} & {\color{black}312.8M} & \textcolor{black}{6h0m} & {\color{black}162.8GFlops} & {\color{black}312.8M} \\
			\end{tabular}
		\end{adjustbox}
	\end{center}
\end{table}

\subsubsection{\textcolor{black}{Gradually Masking cloth-Related Attributes}}
\textcolor{black}{In this chapter, we gradually mask cloth-related attributes to validate our motivation. To prevent any specific attribute from influencing person re-identification results, we randomly mask these attributes at 30\%, 60\%, and 90\%, progressing up to 100\%, as depicted in Table \ref{tab:table5}. Experiments are conducted in the cloth-changing setting of PRCC and LTCC datasets. When 60\% of the cloth-related attributes are masked, compared to 30\%, the rank-1 of PRCC increases by 0.9\%, and the rank-1 of LTCC increases by 1.3\%. When 90\% of the cloth-related attributes are masked, the mAP of PRCC increases by 2.9\%, and the mAP of LTCC increases by 0.3\%. When 100\% of the cloth-related attributes are masked, the rank-1 of PRCC increases by 4.0\%, and the rank-1 of LTCC increases by 2.3\%. As the cloth-related attributes are gradually masked, the accuracy of re-identification improves. The re-identification accuracy remains relatively high even when the cloth-related attributes are partially masked. This suggests that attributes related to clothes affect model re-identification results, and their removal forces the model to learn cloth-insensitive features.}

\begin{table}
    \caption{\textcolor{black}{The experimental results of progressively masking pedestrian cloth-related attributes in the cloth-changing setting of MADE on PRCC and LTCC.}}
    \label{tab:table5}
    \normalsize
    \begin{center}
        \begin{adjustbox}{width=0.7\linewidth}
            \begin{tabular}{c|cc|cc}
                \hline
                \multirow{3}{*}[1ex]{\textcolor{black}{masking ratio}}  & \multicolumn{2}{c|}{\textcolor{black}{PRCC}}  & \multicolumn{2}{c}{\textcolor{black}{LTCC}} \\
                \cline{2-5} 
                & \textcolor{black}{rank-1} & \textcolor{black}{mAP} & \textcolor{black}{rank-1} & \textcolor{black}{mAP} \\
                \hline
                \textcolor{black}{30\%} & \textcolor{black}{62.2} & \textcolor{black}{59.4} & \textcolor{black}{43.6} & \textcolor{black}{22.1} \\
                \textcolor{black}{60\%} & \textcolor{black}{63.1} & \textcolor{black}{59.4} & \textcolor{black}{44.9} & \textcolor{black}{21.4} \\
                \textcolor{black}{90\%} & \textcolor{black}{66.1} & \textcolor{black}{62.3} & \textcolor{black}{44.9} & \textcolor{black}{22.4} \\
                \textcolor{black}{100\%} & \textcolor{black}{66.2} & \textcolor{black}{63.0} & \textcolor{black}{45.9} & \textcolor{black}{25.0} \\
            \end{tabular}
        \end{adjustbox}
    \end{center}
\end{table}

\subsubsection{Number of Stages and TrV Blocks}
\label{stnum}

In this section, we discuss the layering situation of EVA02-large, conduct experiments with MADE, and fuse attribute description features that remove cloth information with low-level to high-level features of images. We tried several stratification scenarios based on experience and summarized the experimental results on the cloth-changing setting of PRCC in Table \ref{tab:table4}. We can observe that the best results are achieved when the number of stages is three, and the number of layers is 10, 10, and 4, respectively. 

\begin{table}
	\caption{{Ablation studies of the different number of stages and TrV blocks for MADE in cloth-changing setting on PRCC. Bold numbers the top score.}}
	\label{tab:table4}
	\normalsize
	\begin{center}
		\begin{adjustbox}{width=0.7\linewidth}
			\begin{tabular}{c|c|cc}
				\hline
				\multirow{3}{*}[1ex]{\# Stage}  & \multirow{3}{*}[1ex]{\# Layer (24)}  & \multicolumn{2}{c}{PRCC}  \\
				\cline{3-4} 
				& & rank-1 & mAP   \\
				\hline
				2 & [12, 12] & 57.6 & 51.5  \\
				2 & [10, 14] & 48.3 & 45.7 \\
				3 & [8, 8, 8] & 63.7 & 57.6  \\
				3 & [8, 12, 4] & 64.4 & 60.0 \\
				3 & [10, 10, 4] & \textbf{66.2} & \textbf{63.0} \\
				4 & [6, 6, 6, 6] & 54.0 & 49.0 \\
				4 & [4, 4, 8, 8] & 54.5 & 51.7 \\
			\end{tabular}
		\end{adjustbox}
	\end{center}
\end{table}

\section{Conclusion}
\label{sec:con}
We propose the Masked Attribute Description Embedding (MADE) method, which integrates a person's visual appearance with attribute description in CC-ReID. The modeling of volatile cloth-sensitive information, including color and type, is challenging and not conducive to identifying persons in CC-ReID. To address this, we introduce multi-modal attribute description information in CC-ReID, which is more obvious and easier to extract and edit than skeletons or contours in original images. We extract descriptions suitable for personal images using an attribute detection model, mask the variable \textcolor{black}{cloth} and color information, and embed it into the image features, compelling the model to discard cloth information. Subsequently, MADE connects and embeds the masked attribute description features encoded by Linear Projection into Transformer blocks at different levels, fusing them with low-level to high-level features of the image. By mapping different feature spaces to a shared latent space, attribute description can be fused with image features, enabling the model to capture the associated information between images and descriptions effectively. We conducted experiments on PRCC, LTCC, Celeb-reID-light, and LaST. Extensive experiments have demonstrated that MADE can effectively utilize personal description information to improve the performance of cloth-changing person re-identification and performs well compared to state-of-the-art methods.

In the future, we intend to expliot the Large Language Models (LLM) to generate better attribute descriptions, which could help futher improve the generalization ability of our method for cloth-changing person re-identification. We will also explore the possibility of our masked attribute description strategy in other cross-modality person re-identification tasks, such as visible-infrared person ReID, cross-resolution person ReID, and sketch based person ReID, because there are also certain attributes which remain unchanged across the modalities.

\bibliographystyle{IEEEtran}
\bibliography{sample-bibliography}
\end{document}